# Automatically Identifying Morphological Relations in Machine-Readable Dictionaries[1][2]

*Joseph Pentheroudakis*
*josephp@microsoft.com*

*Lucy Vanderwende*
*lucyv@microsoft.com*

Microsoft Research
One Microsoft Way
Redmond, WA 98052-6399
USA

*ABSTRACT*

We describe an automated method for identifying classes of morphologically related words in an on-line dictionary, and for linking individual senses in the derived form to one or more senses in the base form by means of morphological relation attributes. We also present an algorithm for computing a score reflecting the system's certainty in these derivational links; this computation relies on the content of semantic relations associated with each sense, which are extracted automatically by parsing each sense definition and subjecting the parse structure to automated semantic analysis. By processing the entire set of headwords in the dictionary in this fashion we create a large set of directed derivational graphs, which can then be accessed by other components in our broad-coverage NLP system. Spurious or unlikely derivations are not discarded, but are rather added to the dictionary and assigned a negative score; this allows the system to handle non-standard uses of these forms.

**Introduction**

The automatic extraction of semantic knowledge from dictionary definitions in machine-readable dictionaries (MRDs) has been the subject of intense research for several years (Amsler 1980, Chodorow et al. 1985, Markowitz et al. 1986, Jensen and Binot 1987, Ahlswede and Evens 1988, Neff and Boguraev 1989, Boguraev and Briscoe 1989, Montemagni and Vanderwende 1992, Dolan et al. 1993). Comparatively little work has been reported, however, on the automatic identification of derivationally related words in MRDs, although the importance of this information to applications such as parsing, generation, machine translation, spell-checking, thesaurus-building and information retrieval has long been recognized (Markowitz et al. 1986, Calzolari 1988, Carroll and Grover 1988, Boguraev and Briscoe 1989). In this paper we report on a method for automatically identifying morphological relations which link headwords in the on-line version of the Longman Dictionary of Contemporary English (LDOCE). During this process, the lexicon is augmented with explicit links connecting individual senses of bases and derived forms, and a certainty score is computed and associated with each such link.

---
[1] Published in *Proceedings of the Ninth Annual Conference of the UW Centre for the New OED and Text Research,* 1993, p. 114-131
[2] We would like to extend our thanks to the other members of the Microsoft NLP group: Bill Dolan, George Heidorn, Karen Jensen, Diana Peterson, Steve Richardson and Eric Ringger.

The method consists of first applying our NLP system's finite-state morphological analyzer to every headword in the dictionary. If a headword is identified as possibly be derived from one or more base words, a separate component, dubbed MORELS, ranks each analysis by comparing the semantic information in the entry of the derived form with the information stored in the putative base form(s). For example, if the noun *conversion* is analyzed as [[*convert*]+*ion*] and [[*converse*]+*ion*], MORELS will apply its scoring algorithm to the semantic content in each sense of *conversion* to that of each sense of *convert* and each sense of *converse*. In this case, the first analysis (*convert*) will receive a high score, while a spurious analysis like the second (*converse*) will receive a low score. Finally, MORELS links each sense of the derived form to the appropriate sense(s) of the base form, and stores the linking information and the associated score in the dictionary as the value of attributes which express derivational relations. This process allows us to systematically identify classes of morphologically related words within LDOCE.

The paper is divided into 5 sections. Section 1 reviews the explicit representation of morphological information in LDOCE. Section 2 presents our method for automatically identifying morphological relations in the entire dictionary. Section 3 discusses the scoring technique. Section 4 discusses our results, Section 5 is a conclusion, and Section 6 describes areas for future work.

## 1. Representation of morphological information in LDOCE

Inflectional information in LDOCE is both explicit and complete. Irregular inflected forms of a word are listed explicitly in LDOCE in the entry for that word, while the remaining forms in the paradigm can be derived by means of rules given in the "Guide to the Dictionary" (p. xxii-xxiv). For example, the entry for the verb *sing* lists the irregular forms *sang* and *sung*, but not the regular forms *sings* and *singing*. On the other hand, a regular form may be included in an entry if it can be problematic for the learner; for example, the entry for the verb *singe* lists the regular past tense form *singed*, presumably to identify this as a valid form of English and to clarify that it is the past tense of *singe,* not *sing*. Finally, if both a regular and an irregular form are possible for a particular slot in the paradigm, both are given; for example, under the verb *dream* we find both *dreamed* and *dreamt*.

In adapting the raw LDOCE data to our own lexical needs, we generated inflectional paradigms automatically drawing on both these irregular forms and the regular rules of English morphology and phonology. Word senses were then marked with indicators representing the appropriate paradigms, again based on the information already present in the data. For example, all senses of the verb *sing* are marked as selecting the SING paradigm. Similarly, the senses of the verb *ring* which inflect like *sing* are also marked as selecting the SING paradigm; however, the sense of the verb *ring* defined as 'to make or be a ring around' is marked as selecting the default paradigm (*rings ringing ringed ringed*). These paradigm markers are then play a crucial role in constraining the morphological processor.

The situation is quite different in the case of derivational morphology, however, where information is generally implicit and scant. Specifically, while it is relatively trivial for human users to generate the entire inflectional paradigm for a word given the information in LDOCE, the same is not true for forms derived from that word. For example, the following words, which are ultimately related to the verb *believe*, are all defined separately in the dictionary (and were automatically identified using the technique described in this paper): *belief*, *believe*, *believable*, *unbelievable*, *believer*, *disbelief*, *disbelieve*, *unbelief*, *unbelieving*, and *unbelievingly*. However, these relations are not marked explicitly in the data, and no level of representation exists where these words are enumerated.

We should mention that there are two classes of exceptions to this lack of overt linking of morphologically related forms. The first class includes undefined derived forms, which are typically appended to the end of the entry of the word from which they are ultimately derived. According to LDOCE, these are words whose meaning "should be clear when the meaning of the suffix is added to the meaning of the base word" (p. xxvii); part of speech and non-predictable grammatical behavior or pronunciation are specified explicitly, and sentences illustrating the use of these forms may also be included. An example is shown in the entry for journalism in Figure 1; the endings for the words *journalistic* and *journalistically* are appended at the end of the entry:

> **jour•na•lism** *n* [U] **1** the work or profession of producing, esp. writing for, JOURNALs (2), esp. newspapers  **2** writing that may be all right for a newspaper, but that lacks imagination and beauty: *His writing is only journalism, not true literature* — **-istic** *adj* — **-istically** *adv* [Wa4]

Figure 1: LDOCE entry for *journalism*

As this example illustrates, the actual derivational history of such undefined forms may be flattened: here, the adjective *journalistic* and the adverb *journalistically* are linked to the noun, even though, morphotactically, the adverb can only be derived from the adjective. Since our initial adaptation of the data preserved the relations among words specified in LDOCE, it suffered from the same flattening; thus, the entries for *journalistic* and *journalistically* both point to the entry for the noun *journalism*.

This example also illustrates the second type of derivational linking, this one due to LDOCE's use of a limited defining vocabulary. If a word is not part of that vocabulary, it is printed in small caps; one such example is the word *journal*, which is printed in small caps in the first sense in the entry for *journalism* above. On occasion, as in this example, this device has the effect of highlighting what is actually the base form of a derived headword. However, it is certainly not true that such base forms will always appear in upper case, or even that they will be present in the definition, just as it is not always the case that all highlighted words in the definition are the base of the word being defined.

Neither type of information about morphological relatedness found in LDOCE can provided us with a systematic picture of the derivational history of a headword. An NLP system, however, which needs to model the knowledge of the speakers of the language, can certainly benefit from such a capability. This information can be introduced manually, but given the size of MRDs and the need for accuracy, completeness and representational flexibility, automatic processing is to be preferred. The technique described in this paper addresses the need for such an automated process.

## 2. Overview

We are not aware of any projects or systems which attempt to identify classes of morphologically related words in an MRD automatically. NLP systems include dictionaries annotated with inflectional and, less often, derivational information (e.g., Russell et al. 1986, Adriaens and Small 1988, Calzolari 1988, among others); it appears that the morphological information in these projects was to a great extent hand-coded. Carroll and Grover (1989) discuss a semi-automated way of generating new entries with associated definitions in the context of dictionary development, and Dumitrescu (1992) describes an integrated environment for an interactive lexicon builder which includes a morphological processing component; still, these methods are in their essence interactive, requiring the assistance of a human user.

Yet a dictionary already contains a great deal of information which can be exploited to identify morphological links automatically. The application of a system's grammar to parse dictionary definitions and acquire semantic knowledge automatically has been described in Jensen and Binot (1987), Montemagni and Vanderwende (1992), and Dolan et al. (1993). Our technique exploits semantic information which has been acquired in this fashion, using it to check whether formally related words are in fact linked by a morphological relationship. The result is a directed graph which connects these words and makes explicit the morphological relations linking them. This allows a user as well as an NLP application access not only to immediate derivatives of a given word, but also to the base form (or forms) of that word and all the other words ultimately derived from it.

In the remainder of this section we will present an overview of the morphological processor and the NLP system in which it is embedded; we will then present the scoring algorithm used by MORELS, and discuss some examples in detail.

### 2.1 The morphological processor

The system's morphological processor relies on morphotactic and allomorphic information stored in morpheme tables. Morphotactic information is expressed in terms of *continuation classes,* in the by now traditional approach advocated by Koskenniemi (1983) and discussed widely in the literature (see esp. Sproat 1992). Specifically, the morphemes that can follow a given morpheme are enumerated in the table for that morpheme. Allomorphic variation, both for prefixes and suffixes, is expressed in a high-level context-free formalism in those tables. Morpheme tables are formally identical to lexical entries; as a result, in addition to information about morphotactics and allomorphy they may include definitions and examples, derived from their definitions in LDOCE.

As an example, consider the entry for the denominal *-er* morpheme (Figure 2), which attaches to nouns to form other nouns (e.g., the pairs *bank/banker*, *geography/geographer* etc.). Lexical entries, and therefore morpheme tables as well, are *records*, or structures of attribute-value pairs; the value of an attribute can be an atom, a list of atoms, another record, or a list of records. By convention, lists of atoms are enclosed in parentheses, and individual records are enclosed in curly brackets.

In morpheme tables, the attribute **Cat** specifies the category of the word formed when this morpheme is attached to another word; in this example the value of **Cat** is *Noun*, indicating that the derived word will be a noun (e.g., the word *geographer*). The attribute **PCat** (*previous category*) gives the category of the words to which the morpheme is allowed to attach; in Figure 2 this is also *Noun*, since the morpheme can attach to nouns only. The value of the attribute **Defin** is the text of the LDOCE definition of this morpheme. The names of the morphemes that can follow it are listed under **NextMorphs**, where the special symbol *None* indicates that it is possible for this morpheme to be followed by no other morphemes (this is one way to avoid having to introduce a zero morpheme for the singular); the only other morpheme allowed in this example is the plural noun morpheme *Noun_Plural*. There is no formal distinction in the system between inflectional and derivational morphemes. Finally, the value of the attribute **Rules** is a list of morphological operations linking derived and base forms. Morpheme records can also be augmented with semantic relation attributes extracted from their definitions, as will be shown below.

When the table is compiled, the affixes are abstracted from each expression, yielding internal representations of the form (_er → _y), (_er → _e) and (_er → _). Using actual examples in the rules makes maintenance of the morpheme tables easy and straightforward (for a similar approach, see Pentheroudakis and Higinbotham (1991)).

```
Cat            Noun
Defin          a person who knows about or works at
Exs            a geographer has studied geography
PCat           Noun
NextMorphs     (Noun_Plural  None)
Rules          geograph er     -> geograph y
               saddl er        -> saddl e
               bank    er      -> bank
```

Figure 2: The denominal *-er* morpheme

We have already defined morpheme tables for the majority of the derivational phenomena listed in Quirk et al. (1985). We are currently expanding the coverage of the morphological processor by consulting works such as Sinclair (1991) as well as Marchand's seminal work on English word-formation (Marchand 1969).

## 2.2  The NLP system

The Microsoft NLP system, of which the morphological processor and MORELS form a part, consists of the following integrated components, each of which applies to the output of the preceding one:

- *a bottom-up, parallel chart parser*, including an algorithm for parse selection and parse recovery;
- *the lexicon*, derived from LDOCE and augmented with semantic relation information extracted automatically from the definitions, as well as with forward and back links to other words and word senses (Dolan et al. 1993);
- *a broad-coverage grammar of English*, consisting of augmented phrase structure rules which produce one or more syntactic analyses for the sentence;
- *a semantic relations processor*, which identifies semantic relations such as *Hypernym*, *Location* etc., in the sense definitions of MRDs (Montemagni and Vanderwende 1992);
- *a modifier attachment component*, which selects the most likely attachment site for modifier phrases; alternative attachment sites will have been clearly marked during syntactic processing;
- a component which computes the *logical form* of sentences, further augmenting the output of the parse with a normalized, abstract representation showing the argument and modifier structure for the sentence;
- *a lexical disambiguation component*, which identifies the most likely sense(s) of polysemous words in the sentence; and
- *an integrated development environment* which includes a high-level language allowing the expression of declarative as well as procedural statements, and tools for on-line dictionary maintenance and rule tracing and debugging.

## 2.3  Extracting semantic relations from dictionary definitions

The technique described in this paper relies crucially on the semantic relations extracted from sense definitions by the semantic relations component. A brief description of that component is presented here. For a detailed description, see Montemagni and Vanderwende (1992) and Dolan et al. (1993).

Sense definitions are first parsed using the system's broad-coverage English grammar; the output of the parse is then subjected to a set of heuristic rules used to identify syntactic and lexical patterns associated with specific relations such as *Hypernym*, *InstrFor* ('instrument for'), etc. Once these patterns are found, the semantic relation attributes are added to the lexical record. For example, the definition of *griddle* is 'a round iron plate which can be used for baking flat cakes (griddle cakes) over a fire'. After the definition is parsed, the semantic relations processor identifies the relations shown in Figure 3 (subordinate relations are indented with respect to superordinate ones):

```
griddle (noun,100) 'a round iron plate which can be used for
                    baking flat cakes (griddle cakes) over a
                    fire'
        Hypernym        {Lemma     "plate"}
        InstrFor        {Lemma     "bake"
                          HasObj   {Lemma    "cake"}
                          LocatedAt {Lemma   "fire"}}
```

Figure 3: Semantic relations extracted from the definition of the noun *griddle*

As can be seen, our approach to semantic representation is primarily lexical and relational. Semantic content is represented in terms of relations between words, a feature which is fully exploited by MORELS.

It bears repeating that semantic relations between words in our lexical database have been identified automatically; with the exception of some of the information in the morphemes, no data have been hand-coded. This property distinguishes our approach from that of other large-scale dictionary or knowledge-base projects such as Wordnet (Miller et al. 1990) or Cyc (Lenat et al. 1989).

## 3. MORELS scoring

The technique used by MORELS in scoring an hypothesized morphological relationship between two words consists of a two-step process. First, a headword in the dictionary is subjected to the system's morphological rules. If a derivational analysis is produced, the semantic relations associated with each sense of the putative base form are compared to those of each sense of the putative derived form. Depending on the result of this comparison process (described in detail below), individual senses of the derived word will be linked to one or more senses of the base word.

### 3.1 The algorithm

This section describes the algorithm used by MORELS to score individual analyses. The algorithm assumes:

- a sense $S$ of a derived word $D$ with a set of semantic relations $SR_d$
- a putative base form $B$ for that word. The semantic relations in the entry for this word are not accessed at this stage; and
- a derivational morpheme $M$ with a set of semantic relations $SR_m$

The algorithm is shown in Figure 4. Comments introduced by a double slash refer to examples discussed in this section.

```
    set score to 0;
    for every semantic relation SR_m in the morpheme M {
        if SR_m is present in the semantic relations SR_d in sense S of the derived word sense D {
            if the lexical content of SR_m matches that of SR_d
                // e.g., 'geographer': 'a person who studies geography'
                increment score by value stored in SR_m;
            else {
                if SR_m is the attribute Hypernym {
                    // e.g., 'banker': 'a player who keeps the bank...'
                    // look up 'player'
                    look up that hypernym in dictionary;
                    obtain its hypernym;
                    compare it to lexical content of SR_m;
                    increment score if successful;
                }
                else {
                    // e.g., cartographer: 'a person who makes maps'
                    // look up 'cartography'
                    look up base word in dictionary;
                    compare its semantic content to that of D;
                    increment score if successful;
                }
            }
        }
        //e.g., 'corner' : semantic relation mismatch; fail.
        else assign score a negative value;
    }
```

Figure 4: MORELS scoring algorithm

Simply put, the algorithm computes a score based on the similarity between the semantic relations in the derived form and those in the morpheme. In effect, we are attempting to establish whether the meaning of a derived word, as represented in its semantic relation attributes and its links to other words in the dictionary, is indeed "the meaning of the suffix [..] added to the meaning of the base word", as suggested in the front matter in LDOCE (p. xxvii).

The next section contains examples of this computation.

### 3.2  Semantic relations in the -*er* morpheme

The lexical entry for the denominal -*er* morpheme, augmented with appropriate semantic relations, is shown in Figure 5; the semantic relation attributes are shown in boldface. In the case of morphemes, semantic relations are extracted automatically from their definitions in LDOCE, but some structure normalization is performed by hand. The value of the attribute *Morels*, also hand-coded, is the amount by which the score will be incremented if that relation is instantiated in the entry for the putative derived word.

```
        Cat        Noun
        Defin      a person who knows about or works at
        Exs        a geographer has studied geography
        PCat       Noun
        NextMorphs (Noun_Plural  None)
        Rules      geograph er    -> geograph y
                   saddl er       -> saddl e
                   bank er        -> bank
    Hypernym       {Lemmas (person)
                    Morels 2}
    SubjOf         {Lemmas (know work)
                    Morels 2
                    HasObj {Morels 10}}
```

Figure 5: The denominal *-er* morpheme augmented with semantic relations

### Example 1: '*geographer*'

The first example, the word *geographer*, is a rather straightforward one. The morphological processor analyzes it as [[*geography*]+*er*], where the *-er* morpheme is the one shown in Figure 5 above. The definition and semantic relations for *geographer* are shown in Figure 6:

```
    geographer (noun,1) 'a person who studies and knows
                        about geography'
              Hypernym  {Lemma      "person"}
              SubjOf    {Lemma      "study"
                         HasObj    {Lemma "geography"}}
                        {Lemma      "know"
                         HasObj    {Lemma "geography"}}
```

Figure 6: Definition and semantic relations for the noun *geographer*

The attributes *Hypernym* and *SubjOf* are instantiated for both the morpheme *-er* and for this sense, as is the nested attribute *HasObj*. Moreover, these sets of semantic relations are structurally isomorphic, and the lemma values for each attribute are identical in the two entries. These two matches contribute 2 points each to the score. Since the base form *geography* appears in the position marked with a *Morels* value of 10 in the morpheme record, the resulting score is 14. In other words, MORELS verifies that the existing definition and semantic relations for the noun *geographer* as given in the dictionary are indeed a compositional function of the word *geography* and the *-er* morpheme.

### Example 2: '*cartographer*'

Not all definitions reflect the compositional nature of the semantic relations in such a straightforward manner. For example, consider the noun *cartographer*, shown in Figure 7, which the morphological processor analyzes as [[*cartography*]+*er*]:

```
cartographer (noun,1) 'a person who makes maps; map-maker'
            Hypernym    {Lemma    "person"}
            SubjOf      {Lemma    "make"
                         HasObj   {Lemma    "map"}}
            Synonym     (map-maker)
```

Figure 7: Definition and semantic relations for the noun *cartographer*

There is no direct lexical match between the base word *cartography* and the value of the *HasObj* attribute in the morpheme map. If we were to limit ourselves to a simple comparison of lexical forms in the records being compared, we would have to assign a low score (a score of 2) to this derivation. However, in such cases MORELS goes on to look up the definition of the base form *cartography* to determine whether its semantic content matches that in the derived form. The definition and semantic relations for the noun *cartography* are shown in Figure 8:

```
cartography (noun,1) 'the science or art of making maps'
      Hypernym    {Lemma       "make"
                   Classifier   {Lemma    "art"}
                                {Lemma    "science"}
                   HasObj       {Lemma    "map"}}}
```

Figure 8: Definition and semantic relations for the noun *cartography*

As can be seen, the *SubjOf* attribute in *cartographer* matches the *Hypernym* information in the base form *cartography*. Note that we have, in this process, discovered and confirmed a link between two words in the dictionary which was not present in the original LDOCE data; this link will subsequently be stored in the dictionary, as shown below.

**Example 3: '*banker*'**

Now consider the definition and semantic relations in the second sense of the noun *banker*:

```
banker (noun,2) 'the player who keeps the bank in various
              games of chance'
       Hypernym    {Lemma        "player"}
       SubjOf      {Lemma        "keep"
                    HasObj       {Lemma    "bank"}
                    LocatedAt    {Lemma    "game"}}
```

Figure 9: Definition and semantic relations for sense 2 of *banker*

When MORELS compares the lexical content of the Hypernym attribute in this record with that in the morpheme, the word *player* will not match *person*. Continuing to search for a way to link the two senses, MORELS will look up the word *player* in the dictionary and check its hypernym; and in fact, one of the hypernyms of *player* in the dictionary is the word *person*, thus allowing MORELS to count the use of the word *player* here as a hit.

Note that, since the verb *keep* does not match either *know* or *study* in the *SubjOf* attribute for *-er*, this derivation will receive as lower score than the derivations in the previous two examples. The score thus reflects the fact that this sense of *banker* is not as straightforward a composition of the base and the morpheme *-er*.

**Example 4: '*corner*'**

Finally, consider the case of the word *corner*, which is analyzed by the morphological processor as [[*corn*]+*er*]. None of the noun senses for this word in LDOCE corresponds to the regular compositional meaning that we might expect it to have: *'a person who knows about *corn* or works at *corn*'. As a result, MORELS will fail, returning a negative value. Even in this case, however, a link will be established between the word *corn* and the word *corner*, but not between any of their existing senses. This link will have a negative value, which indicates that, while this pairing is formally regular, it does not appear to represent a straightforward case of derivation.

**3.3 Linking to base senses**

So far we have described how MORELS checks whether the definition of a derived form supports a compositional analysis of that word. In this section we consider another important task performed by MORELS: attempting to link individual senses in each derived word with one or more specific senses in the base word. To achieve this, a set of heuristic rules which examines each hypothesized base/derived form pair, attempting to discover structural and semantic similarities between each pair of senses. The resulting link score reflects the similarity between a pair of base and derived form senses, and it is associated with the link to each sense in the base form, as will be shown below.

As an example, consider the word *conversion*. As was pointed out earlier, the morphological processor suggests two base forms for this word, the verbs *convert* and *converse*. The challenge is to link individual senses of the noun *conversion* to individual senses of the verb *convert*, and to represent the fact that none of the senses of the noun *conversion* can be plausibly linked to the verb *converse*. The definitions for the noun *conversion* are shown in Figure 10; for the sake of brevity, the semantic relations extracted from the definitions are not included. The number following the part-of-speech designation reflects our internal, normalized numbering of LDOCE senses; the order of the definitions, however, follows that of the printed version.

```
conversion (noun,100): 'the act of converting'
conversion (noun,101): 'a change from one use or purpose to another'
conversion (noun,102): 'a change in which a person accepts completely a
                       new religion, political belief, etc.'
conversion (noun,103): '(in rugby and American football) (an example of)
                       the act of kicking the ball over the bar of the
                       goalposts'
```

Figure 10: Definitions for the noun *conversion*

The definitions for the verbs *convert* and *converse* are shown in Figures 11 and 12, respectively:

```
convert (verb,100): 'to persuade a person to accept a particular
                    religion, political belief, etc.'
convert (verb,101): 'to change one's religion'
convert (verb,102): 'to (cause to) change to or into another form,
                    substance, or state, or from one use or purpose to
                    another'
convert (verb,103): '(of one type of money) to change into another type
                    of money of equal value'
convert (verb,104): 'to cause (one type of money) to change into another
                    of equal value'
convert (verb,105): '(in rugby and American football) to kick (a ball)
                    over the bar of the goalposts'
```

Figure 11: Definitions for the verb *convert*

```
converse (verb,100): 'to talk informally'
```

Figure 12: Definition for the verb *converse*

During the first pass, described in the previous section, the analysis of *conversion* as [[*convert*]+*ion*] receives a high score, while the alternative analysis, [[*converse*]+*ion*], receives a negative score. If the score is greater than zero, MORELS performs a pairwise comparison of the semantic relations in each of the senses shown above; in this case, it will provide a link score for each sense of *conversion* linked to each sense of *convert*. The link score is incremented for each sense using the following heuristics:

- *Intersection in syntactic marking*. Subcategorized prepositions are often marked on a verb sense and on its derived nominalization(s). Increment link score for every preposition that matches, decrement if there is no match.
- *Intersection in domain attribute*. One of the semantic relations extracted from the definition text is the attribute *Domain*; this corresponds to material like the

    definition-initial phrase '(in rugby and American football)' in *conversion (noun, 103)* and *convert (verb,105)*. Increment link score for every match, decrement if there is no match.
- *Compare the content of the semantic relations* in the derived senses to that of the *Hypernym* attribute in the base. Increment link score for every match on the word or its hypernym.

Once a link score has been computed for a pair of derived and base sense records, MORELS is ready to update the lexicon with the result of its computation. A symmetrical pair of morphological attributes is associated with each morpheme; for example, the *-er* morpheme discussed earlier is associated with the attributes *Profsn* ('profession') and *ProfsnOf* ('profession of'), while the nominalizing morpheme *-ion* is associated with the attributes *Nomnlz* and *NomnlzOf*. The first attribute in each pair is added to the base sense(s), while the second will be introduced to the senses of the derived form. The value of each attribute is a list of records specifying a headword to which the sense is linked, the particular sense number in that entry, and the score associated with the link. For example, the noun *cartography* will contain a *Profsn* attribute pointing to the word *cartographer*; the latter, in turn, will contain a *ProfsnOf* attribute pointing to *cartography*. The presence of this attributes allows us to represent derivational paradigms as directed graphs linking morphologically related words.

For a relatively complex example, consider the updated record for sense 102 of the noun *conversion* (Figure 13); our internal sense numbers appear as the value of the *Ldoce* attribute, and they correspond to the numbers given in Figures 10 and 11 above. The sense of *convert* which is assigned the highest-scoring *NomnlzOf* link by MORELS is the second sense in LDOCE (sense 101 in our normalized numbering), 'to change one's religion'. In the current implementation, all non-zero links to senses of *convert* are shown.

```
    Ldoce   102
    Cat     Noun
    Defin   "a change in which a person accepts
            completely a new religion, political belief, etc."
    Hypernym
            {Lemma      "change"}
            {Lemma      "accept"
             HasObj     {Lemma      "religion"}
                        {Lemma      "belief"}
                        {Lemma      "?"}}
    Manner
            {Lemma      "completely"}
    NomnlzOf
            {Ldoce      100
             Lemma      "convert"
             Morels      5}

            {Ldoce      101
             Lemma      "convert"
             Morels     20}

            {Ldoce      102
             Lemma      "convert"
             Morels      5}

            {Ldoce      103
             Lemma      "convert"
             Morels     15}

            {Ldoce      104
             Lemma      "convert"
             Morels     15}
```

Figure 13: Sense 102 of the noun *conversion* after computing links to senses of *convert*

### 4. Conclusion

We have described an automated method for identifying classes of morphologically related words in an on-line dictionary, and for linking individual senses in the derived form to one or more senses in the base form by means of morphological relation attributes. We have also presented an algorithm for computing a score reflecting the system's certainty in these derivational links; this computation relies on the content of semantic relations associated with each sense, which are extracted automatically by parsing each sense definition and subjecting the parse structure to automated semantic analysis. By processing the entire set of headwords in the dictionary in this fashion we create a large set of directed derivational graphs, which can then be accessed by other components in our broad-coverage NLP system.

## 5. Results

### 5.1 Morphological processing

We report first on the performance of the morphological processor, since our method relies crucially on the analyses hypothesized by this component.

Counting different parts of speech for the same word as different words, our version of LDOCE contains 45,808 words. The initial phase of morphological processing, which does not have access to semantic information, currently identifies 13,028 forms as being derived from other forms; of these, 1,748 forms have ambiguous derivations (e.g., pairs like *deduction/deduce* and *deduction/deduct*). Running the processor on all 45,808 forms takes just over three hours on a 486/66 PC.

As was mentioned in section 2, LDOCE does occasionally list derivationally related forms as undefined words at the end of an entry. The total number of such words linked to LDOCE base words is 5,390 (that is, there is a total of 5,390 undefined forms in the dictionary). Our analyzer agreed with LDOCE on 4,798 of these forms, or 89% of the time. There are 308 derivations in LDOCE which suffer from the type of flattening described in section 2, and for which our analyzer produced the correct structure; if we consider these as successful, our agreement ratio with LDOCE becomes 95%.

We hand-checked a random sample of 340 words for which the analyzer returned a polymorphemic analysis. We identified 19 spurious analyses, for an accuracy ration of 94.4%. Using common statistical techniques we estimate that this 94.4% precision rate is representative of the entire set of analyses, with a margin of error of ± 2.5%. To estimate recall, we hand-checked a sample of about 800 words. We found that 283 were morphologically complex, and of those the analyzer failed to analyze 28. We estimate that this 90% success rate is representative of the entire set of analyses, with a margin of error of ± 3%.

A list of affixes matched is given in Table 1; these are sorted by frequency of occurrence.

| affix | count | affix | count |
|---|---|---|---|
| -ly | 2992 | in- | 218 |
| -ness | 1330 | -ive ('creative') | 201 |
| -ation | 1160 | un- | 189 |
| -er ('reader') | 1123 | -less | 177 |
| -er ('banker') | 1032 | de- | 174 |
| -ity | 552 | -al | 169 |
| -y ('creamy') | 468 | re- | 162 |
| -able/-ible | 320 | -ous | 157 |
| -ize | 320 | -ful | 100 |
| -ence | 301 | pre- | 99 |
| -ist ('anarchist') | 285 | -ship | 72 |
| -ism | 275 | inter- | 67 |
| dis- | 231 | over- | 65 |
| -ment | 225 | -ify | 59 |
| -ess ('actress') | 51 | -hood | 19 |
| mis- | 49 | -dom | 17 |
| -ery | 45 | post- | 17 |
| -en | 34 | -ent | 16 |
| counter- | 32 | -wise | 13 |
| -y ('difficulty') | 32 | -ate ('authenticate') | 11 |
| -ee | 29 | after- | 10 |
| -ish | 23 | -worthy | 8 |
| under- | 23 | -wise | 5 |
| non- | 22 | multi- | 4 |
| anti- | 21 | -ese | 3 |
| -like | 20 | vice- | 1 |

Table 1: Distribution of some affixes in LDOCE headwords

## 5.2 MORELS

Next we look at the performance of MORELS, the component which evaluates the analyses and establishes the links between related base/derived word senses.

To evaluate the output of MORELS, we generated a file of tuples of the form (*derived word*, *part of speech*, *derived sense number*, *base word*, *part of speech*, *base sense number*, *score*). Consider the word *viewer*, which formally admits the deverbal analysis [[*view*$_{\text{verb}}$]+*er*] and the denominal analysis [[*view*$_{\text{noun}}$]+*er*]. The section of the file that contains the output for sense 100 of the word *viewer*, 'a person watching television', shows that it is linked to all noun senses of the headword *view* by a score of -4, indicating a possibly spurious analysis, while it is linked to verb sense 117, 'to watch (esp. television)' by a high score of 26:

```
viewer, noun, 100, view, noun, 103, -4
viewer, noun, 100, view, noun, 104, -4
viewer, noun, 100, view, noun, 105, -4
viewer, noun, 100, view, noun, 106, -4
viewer, noun, 100, view, noun, 107, -4
viewer, noun, 100, view, noun, 108, -4
viewer, noun, 100, view, verb, 117, -4
viewer, noun, 100, view, verb, 118, -4
viewer, noun, 100, view, verb, 119, 26
```

Table 2: MORELS output for the word *viewer*

Of all the proposed deverbal analyses in the corpus of 245 words ending in -er, MORELS identified an appropriate target sense in the base headword entry 65% of the time. MORELS did correctly identify several spurious analyses, for example *[[*flow*]+*er*] for *flower* and *[[*show*]+*er*] for *shower*. In general, however, its success rate is still relatively low, both in terms of precision and recall. Since MORELS relies heavily on the output of the parse and the semantic relations processor, we expect its success rate to improve dramatically as the output of those components continues to improve. Additionally, we expect the performance of MORELS to improve as we continue to refine the semantic content of the morphemes to allow us a better match with the semantic relations in the word sense definitions.

## 6. Future work

The discussion in this paper is limited to our lexicon, which was derived from LDOCE. The method is not limited to a particular dictionary, however; we are in fact planning to apply it to data from the American Heritage Dictionary (Third Edition), and we will be reporting on those results in the future.

One of the further areas of research we will pursue is to allow MORELS to modify its score based on information from other parts of dictionary entries, such as, for example, from the pronunciation or possibly even the etymology fields. For instance, the morphological processor currently proposes two analyses for the word *cellist*, namely [[*cello*]+*ist*] and [[*cell*]+*ist*]. Even though MORELS correctly assigns the latter a negative link score, using the pronunciation information would provide further evidence against this analysis.

We are also interested in the automatic identification of suppletive paradigms such as *king/royal*, *dog/canine*, etc. We can apply our method to identify such derivational paradigms, and it certainly seems desirable to ultimately link the noun *king* and the adjective *royal* with the same denominal adjective relation that is used to link formally related noun/adjective pairs, such as *culture/cultural*, *history/historic* etc. MORELS can also be used to identify semantic relation patterns in the definitions of denominal adjectives, for example, and apply this knowledge to determining suppletive noun/adjective pairs, in a manner reminiscent of the work described in Markowitz et al. (1986).